%% file: EmbryoDiff.tex
\documentclass[letterpaper]{article} 
\usepackage{aaai2026}  
\usepackage{times}  
\usepackage{helvet}  
\usepackage{courier}  
\usepackage[hyphens]{url}  
\usepackage{graphicx} 
\usepackage{natbib}  
\usepackage{caption} 
\usepackage{algorithm}
\usepackage{algorithmic}
\usepackage{amssymb}
\usepackage{amsmath}
\usepackage{multirow}
\usepackage{booktabs}
\usepackage{rotating}
\usepackage{array}

\usepackage{newfloat}
\usepackage{listings}
\DeclareCaptionStyle{ruled}{labelfont=normalfont,labelsep=colon,strut=off} 
\floatstyle{ruled}
\newfloat{listing}{tb}{lst}{}
\floatname{listing}{Listing}
\title{EmbryoDiff: A Conditional Diffusion Framework with Multi-Focal Feature Fusion for Fine-Grained Embryo Developmental Stage Recognition}
\author{
    Yong Sun,
    Zhengjie Zhang,
    Junyu Shi,
    Zhiyuan Zhang,
    Lijiang Liu,
    Qiang Nie\textsuperscript{*}
}
\affiliations{
    The Hong Kong University of Science and Technology (Guangzhou)\\
    qiangnie@hkust-gz.edu.cn
}

\usepackage{bibentry}

\begin{document}

\maketitle

\begin{abstract}
Identification of fine-grained embryo developmental stages during In Vitro Fertilization (IVF) is crucial for assessing embryo viability. Although recent deep learning methods have achieved promising accuracy, existing discriminative models fail to utilize the distributional prior of embryonic development to improve accuracy. Moreover, their reliance on single-focal information leads to incomplete embryonic representations, making them susceptible to feature ambiguity under cell occlusions. To address these limitations, we propose EmbryoDiff, a two-stage diffusion-based framework that formulates the task as a conditional sequence denoising process. Specifically, we first train and freeze a frame-level encoder to extract robust multi-focal features. In the second stage, we introduce a Multi-Focal Feature Fusion Strategy that aggregates information across focal planes to construct a 3D-aware morphological representation, effectively alleviating ambiguities arising from cell occlusions. Building on this fused representation, we derive complementary semantic and boundary cues and design a Hybrid Semantic-Boundary Condition Block to inject them into the diffusion-based denoising process, enabling accurate embryonic stage classification. Extensive experiments on two benchmark datasets show that our method achieves state-of-the-art results. Notably, with only a single denoising step, our model obtains the best average test performance, reaching 82.8\% and 81.3\% accuracy on the two datasets, respectively.
\end{abstract}

\begin{links}
    \link{Code}{https://github.com/RIL-Lab/EmbryoDiff}
\end{links}

\begin{figure}[t]
\centering
\includegraphics[width=0.98\columnwidth]{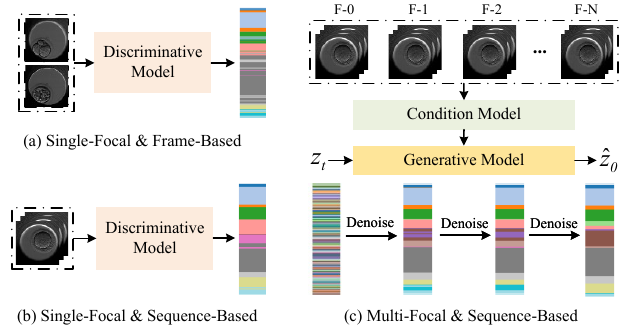} 
\caption{Comparison of different frameworks. (a) Frame-based methods process frames independently. (b) Sequence-based methods capture temporal dependencies from single-view videos. (c) Our method integrates multi-focal TLM videos with a conditional generative model. Different colors indicate different developmental stages.}
\label{pipcom}
\end{figure}

\section{Introduction}

Infertility has affected approximately 80 million individuals of reproductive age worldwide \cite{inhorn2015infertility}. In response, in vitro fertilization (IVF) has become a central intervention. In current clinical practice, embryologists examine Time-Lapse Monitoring (TLM) videos of in vitro fertilized embryos to extract both morphological and morphokinetic features, selecting the most viable embryo for transfer, which is labor-intensive and time-consuming. Recently, deep learning has been adopted in assisted reproductive technology, with applications spanning blastocyst evaluation \cite{blastocysts_eva1, blastocysts_eva2}, embryo grading \cite{grading}, and viability prediction \cite{viability_1, viability_2, viability_3}.

A crucial subtask in embryo evaluation is embryo developmental stage classification. The goal is to assign a developmental phase to each frame in a TLM sequence, thereby offering morphokinetic cues that are essential for embryo quality assessment~\cite{task_1, task_2}. In recent years, considerable effort has been devoted to automating this labor-intensive and expert-dependent process using deep learning techniques. As illustrated in Figure~\ref{pipcom}(a) and (b), existing approaches can be broadly categorized into two groups: (1) \textbf{Frame-based methods}~\cite{frame_DTLEmbryo, frame_inception, frame_emca, frame_cascaded}, which process each frame independently and often suffer from limited temporal coherence due to the absence of temporal modeling; and (2) \textbf{Sequence-based methods}~\cite{seq_cnn_crf, seq_latefusion, seq_synergic, seq_canat2024novel}, which take entire TLM sequences as input and explicitly model temporal dependencies among developmental stages. Despite the progress achieved, accurate fine-grained developmental stage classification remains challenging due to several key limitations.

First, existing discriminative methods fail to leverage the inherent distributional priors of embryonic development. Although individual embryos vary in developmental timing, their overall trajectories generally follow well-established biological principles, such as monotonic stage transitions, characteristic phase durations, and the coordinated occurrence of developmental events. Discriminative models, however, map visual inputs directly to stage labels without modeling these underlying priors. Consequently, even state-of-the-art approaches are prone to unstable or unreliable predictions under challenging conditions.

Moreover, prior methods suffer from incomplete embryonic morphological representation due to their reliance on single-focal-plane TLM videos, which capture only a partial 2D projection of the embryo’s inherently 3D structure. As development progresses through successive cell divisions, the increasing number of blastomeres leads to frequent cell occlusions in these 2D projections. In clinical practice, embryologists mitigate this issue by integrating visual cues across multiple focal planes to reconstruct a more complete 3D morphological context. In contrast, current approaches are restricted to a single central focal plane and therefore lack the contextual information required to disambiguate overlapping cells in densely packed configurations.

Furthermore, model performance is constrained by data quality and evaluation practices. Noisy annotations in the public dataset \cite{dataset_problem, dataset_gomez} compromise training stability, and frame-level metrics ignore the sequence nature of embryo developmental progression.

In this paper, we propose \textbf{EmbryoDiff}, a two-stage diffusion-based framework (Figure~\ref{pipcom}(c)) designed to address these challenges. Unlike discriminative methods, diffusion models learn the underlying data distribution and can therefore naturally encode the developmental priors that govern embryonic progression. Inspired by conditional diffusion architectures such as LDM~\cite{LDM} and DiT~\cite{DiT}, EmbryoDiff formulates the task as a conditional sequence denoising process and uses multi-focal TLM sequences as conditioning inputs for guidance, enabling the recovery of more accurate and biologically plausible developmental stage labels from noisy sequences. To reduce computational burden, we adopt a two-stage training strategy. In the first stage, a frame-based model is trained on individual frames and fixed as a feature extractor. In the second stage, a conditional diffusion model is trained to obtain final classification results. We extract features from multi-focal-plane TLM videos and fuse them via a Multi-Focal Feature Fusion Strategy, constructing more comprehensive, 3D-context-aware morphological representations. To provide comprehensive conditional guidance, we design parallel semantic and boundary condition encoders to capture complementary cues, and introduce a Hybrid Semantic-Boundary Condition Block to effectively inject these features into the denoising process. By leveraging the distributional priors encoded in the diffusion model, enhanced 3D-aware representations, and comprehensive conditional guidance, EmbryoDiff achieves more accurate stage classification and outperforms prior methods.

Beyond algorithmic improvements, we also enhance data quality and evaluation protocols. We systematically correct annotations in a public dataset \cite{dataset_gomez} by removing errors and refining stage boundaries. Inspired by action segmentation \cite{ASFormer, FACT}, we adopt both frame-level and sequence-level metrics to enable a more comprehensive evaluation.

In summary, our contributions are three-fold:
\begin{itemize}
    \item We introduce EmbryoDiff, the first diffusion-based framework for embryonic stage recognition, and the first to leverage multi-focal-plane video inputs for holistic embryo representation.
    \item We enhance the data quality of a public dataset and leverage sequence-level evaluation metrics to complement frame-level accuracy, enabling more reliable training and assessment.
    \item We conduct extensive experiments and ablations, achieving state-of-the-art results on two benchmark datasets and providing insights for future research.
\end{itemize}

\section{Related Works}

\subsubsection{Diffusion Model for Visual Perception.} 
Diffusion models \cite{DDPM, DDIM} generate data samples through an iterative denoising process and have achieved state-of-the-art performance in various generation tasks. Recent studies have shown that, with appropriate conditional guidance, diffusion models can also be effectively applied to visual perception tasks. These models have demonstrated promising results in a wide range of applications, including image segmentation \cite{diff_medsegdiff, diff_ambseg, diff_insseg, ddp}, depth estimation \cite{diff_unwdep, diff_depesti}, object detection \cite{diff_dect1, diff_act2}, action segmentation \cite{diff_act1, diff_act2}, temporal action localization \cite{diff_actloc1} and anomaly detection \cite{diff_ad1, diff_ad2, diff_ad3}. However, in embryo developmental stage recognition, existing methods rely on discriminative models without considering the distribution prior of embryo development. In this paper, we make the first attempt to leverage the diffusion-based framework for this task.

\subsubsection{Embryo Developmental Stage Classification.} Embryo developmental stage classification provides critical morphokinetic features for embryo selection, such as division timing and stage duration. Recent studies have explored deep learning to automate this process. Early methods adopted CNNs like InceptionV3 \cite{frame_inception} and ResNet \cite{dataset_gomez} for frame-wise classification, but ignored temporal dependencies. Later works \cite{seq_canat2024novel, seq_cnn_crf} incorporated temporal modeling, improving performance. Embryosformer \cite{seq_embryosformer} further enhanced accuracy by integrating deformable attention and transformer decoder. However, accurate identification of fine-grained developmental stages remains challenging by soly using single-focal information. To this end, in this paper, we incorporate multi-focal feature fusion to improve the classification performance.

\begin{figure*}
\centering
\includegraphics[width=0.85\textwidth]{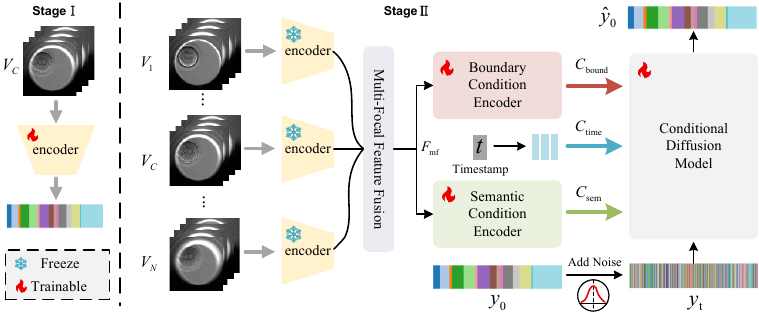} 
\caption{Overview of the two-stage EmbryoDiff framework. Stage 1 (left) trains a frame-based model, while Stage 2 (right) utilizes multi-focal videos to generate semantic and boundary conditions that guide the diffusion-based denoising process.}
\label{framework}
\end{figure*}

\section{Method}
EmbryoDiff consists of two stages (Figure~\ref{framework}). Stage~1 trains a frame-based model on central-plane TLM videos and freezes it as a multi-focal feature extractor. Stage~2 trains the diffusion model: the extracted multi-focal features are fused by the Multi-Focal Feature Fusion Strategy and subsequently encoded by the Boundary Condition Encoder and the Semantic Condition Encoder to generate robust and complementary conditional signals. These signals are then injected into the diffusion model to recover clean stage sequences from noisy inputs.

\subsection{Frame-Based Classification Model}
Multi-focal TLM videos are 5D data and consist of hundreds of frames per sequence, posing huge computational and memory challenges for end-to-end training. To this end, we adopt a two-stage framework. In Stage 1, following prior work \cite{dataset_gomez}, we treat each frame independently and train a frame-based classifier to learn robust visual representations using TLM videos $V_c$ from central focal plane. In practice, we use ResNet-50 \cite{resnet} as the backbone and optimize it with standard cross-entropy loss.

\subsection{Diffusion-Based Classification Model}
In Stage~2, we condition the diffusion model on TLM sequences to denoise and generate coherent developmental stage predictions.

\subsubsection{Training}

Let $\{V_1,\dots,V_N\}$ denote the videos from $N$ focal planes. The pretrained frame-level encoder processes each video to extract its feature sequence $F_i=\mathrm{Enc}(V_i)\in\mathbb{R}^{T\times D}$, where $D=2048$. These features are then fused by the Multi-Focal Feature Fusion Strategy to form a holistic representation $F_{\text{mf}}=\Phi(F_1,\dots,F_N)$, where $\Phi(\cdot)$ denotes the fusion operation. The fused representation is further encoded by the Semantic Condition Encoder and the Boundary Condition Encoder to obtain the complementaru semantic condition $C_{\text{sem}}$ and boundary condition $C_{\text{bound}}$.

For training the diffusion model, inspired by diffusion-based segmentation methods~\cite{ddp, diff_medsegdiff, diff_act1}, we add noise to ground-truth labels and train the diffusion model to reverse this process and recover clean sequences. Suppose that we have $c$ development stages. We first map the $T$-length one-hot stage labels $\mathbf{y} \in \mathbb{R}^{T \times c}$ into semantic embeddings $\mathbf{y}_0 \in \mathbb{R}^{T \times d}$ using a learnable embedding layer, where $d$ denotes the embedding dimension. We then apply the standard forward diffusion process to corrupt the label sequence with Gaussian noise:
\begin{equation}
    q(\mathbf{y}_t \mid \mathbf{y}_0) = \mathcal{N}(\mathbf{y}_t; \sqrt{\bar{\alpha}_t} \mathbf{y}_0, (1 - \bar{\alpha}_t)\mathbf{I}),
\end{equation}
where $ \bar{\alpha}_t $ is the cumulative noise scaling factor at time step $ t $. To reverse the process, the diffusion model receives the noisy sequence $ \mathbf{y}_t $, the timestep embedding $ C_\text{time} $, and two types of conditional features $ C_\text{sem} $ and $ C_\text{bound} $ to make predictions:
\begin{equation}
    \hat{\mathbf{y}}_0 = f_\theta(\mathbf{y}_t, C_\text{time}, C_\text{sem}, C_\text{bound}),
\end{equation}
where $ f_\theta $ denotes the diffusion model parameterized by $ \theta $. We adopt a DiT-style~\cite{DiT} diffusion decoder, where the timestep embedding $C_\text{time}$ is first added to the noisy features and then processed by stacked Hybrid Semantic-Boundary Condition Blocks to inject the two complementary condition signals into the denoising process. Finally, the predicted embeddings $ \hat{\mathbf{y}}_0 $ are mapped back to the one-hot space to obtain the predicted labels $ \hat{\mathbf{y}} $.

\subsubsection{Inference}

In the inference phase, we sample a random noise from the standard Gaussian distribution.
\begin{equation}
    \mathbf{y}_T \sim \mathcal{N}(\mathbf{0}, \mathbf{I}).
\end{equation}
Then, the DDIM \cite{DDIM} sampling strategy is utilized and we can generate high-quality predictions through several denoising steps.

\subsection{Multi-Focal Feature Fusion}
Embryonic development is a dynamic process of cell division and 3D reorganization, often causing cell occlusions that hinder feature learning \cite{seq_cnn_crf}. To capture a more comprehensive morphology, we integrate features from multiple focal planes. Specifically, we adopt a prefusion strategy, in which features from multi-focal videos are fused before entering the condition encoders. This design brings two key benefits. First, it substantially reduces the computational cost of the encoders by avoiding attention over excessively long feature sequences. Second, it improves the framework’s flexibility, allowing it to seamlessly accommodate single-focal settings by simply disabling the fusion step without modifying any other components. In practice, we employ a simple average pooling operation along the focal-plane dimension to obtain the fused feature $F_{\text{mf}} \in \mathbb{R}^{T \times D}$ as:
\begin{equation}
    F_{\text{mf}} = \frac{1}{N} \sum_{i=1}^{N} F_i,
\end{equation}
This strategy shifts from 2D in-plane analysis to a 3D-aware modeling of embryonic structure, capturing spatial context that enhances feature robustness against cell occlusions.

\begin{figure}[t]
\centering
\includegraphics[width=0.90\columnwidth]{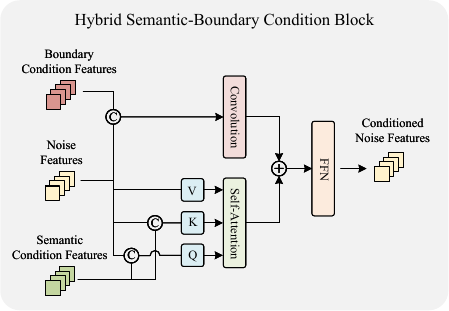} 
\caption{The architecture of the Hybrid Semantic-Boundary Condition Block. For clarity, the residual connection between the original and conditioned noise features is omitted.}
\label{condition_block}
\end{figure}

\subsection{Complementary Semantic-Boundary Conditions}

The performance of conditional diffusion models is highly dependent on the quality of conditioning signals. Prior work MedSefDiff-V2 \cite{diff_medsegdiff} uses an auxiliary segmentation loss to train the condition network for semantic guidance in denoising. However, such semantic cues alone provide limited discriminative power for embryo stage recognition, particularly in transitional phases where morphological changes are subtle and stage boundaries are difficult to distinguish. To this end, we introduce complementary boundary-aware condition features that work in conjunction with semantic cues to provide more discriminative guidance during the denoising process. As shown in Figure~\ref{framework}, a dual-branch encoder extracts semantic features $C_\text{sem}$ and boundary features $C_\text{bound}$ from the fused multi-focal sequence features $F_{\text{mf}}$. Both branches adopt the ASFormer architecture~\cite{ASFormer}, capturing local and global temporal dependencies. 

\subsubsection{Hybrid Semantic-Boundary Condition Block}

As shown in Figure~\ref{condition_block}, each block handles noise and condition features via two complementary pathways. In the boundary-aware path, boundary condition features are concatenated with noise features and processed by a convolution layer to capture local interactions. In the semantic-aware path, semantic condition features are concatenated with noise features to form the query-key pairs, while the noise features serve as values in a self-attention module to inject long-range semantic context. The outputs from both pathways are summed and passed through a Feed-Forward Network (FFN) to produce the final conditioned noise features.

This hybrid design ensures the synergistic integration of local boundary cues and global semantic context, assisting the denoising process to generate both semantic correct and boundary-accurate predictions. 

\subsection{Loss Functions}
Multiple losses are jointly optimized in Stage~2. The semantic branch uses a stage classification loss $\mathcal{L}_{\text{sem}}$ and a temporal smoothing loss $\mathcal{L}_{\text{smooth}}$ ~\cite{ASFormer}, while the boundary branch uses a binary boundary classification loss $\mathcal{L}_{\text{bound}}$. The diffusion model is also optimized with a stage classification loss $\mathcal{L}_{\text{diff}}$. The overall objective is:
\begin{equation}
\mathcal{L} = \lambda_1 \mathcal{L}_{\text{sem}} + \lambda_2 \mathcal{L}_{\text{smooth}} + \lambda_3 \mathcal{L}_{\text{bound}} + \lambda_4 \mathcal{L}_{\text{diff}},
\end{equation}
All classification losses are standard cross-entropy losses. We set $\lambda_1{=}0.8$, $\lambda_2{=}0.3$, $\lambda_3{=}0.5$, and $\lambda_4{=}1.0$ to balance their contributions.

\begin{table*}[htbp]
  \centering
    \begin{tabular}{
      p{1em}<{\centering}       
      p{3em}<{\centering}|      
      p{2.1em}<{\centering}   
      p{2.1em}<{\centering}   
      p{7em}<{\centering}   
      p{2.1em}<{\centering}|   
      p{2.1em}<{\centering}   
      p{2.1em}<{\centering}   
      p{7em}<{\centering}  
      p{2.1em}<{\centering}    
    }
    \toprule
    \multicolumn{2}{c|}{\multirow{2}[3]{*}{Methods}} & \multicolumn{4}{c|}{Multi-Focal Human Embryos Dataset} & \multicolumn{4}{c}{Single-Focal Human Embryos Dataset} \\
\cmidrule{3-10}    \multicolumn{2}{c|}{} & Acc & Edit & F1@\{10, 25, 50\} & Avg & Acc  & Edit  & F1@\{10, 25, 50\} & Avg \\
    \midrule
    \multicolumn{1}{c|}{\multirow{3}{*}{\begin{turn}{-90}Frame\end{turn}}} & \multicolumn{1}{c|}{ResNet} & 74.3 & 24.7 & 30.3 / 26.3 /19.1 & 34.9  & 75.8 & 24.1 & 30.4 / 26.1 / 18.9 & 35.1 \\
    \multicolumn{1}{c|}{} & \multicolumn{1}{c|}{DLTEmbryo} & 69.2 & 24.3 & 28.4 / 24.3 / 17.8 & 32.8 & 71.3 & 22.5 & 27.3 / 22.7 / 15.8 & 31.9 \\
    \multicolumn{1}{c|}{} & \multicolumn{1}{c|}{InceptionV3} & 76.2 & 29.8 & 34.7 / 30.9 / 23.9 & 39.1 & 75.9 & 27.0 & 34.1 / 29.5 / 22.1 & 37.7 \\
    \midrule
    \multicolumn{1}{c|}{\multirow{5}{*}{\begin{turn}{-90}Sequence\end{turn}}} & \multicolumn{1}{c|}{ResNet-LSTM} & 75.0 & 46.5 & 51.2 / 45.7 / 35.7 & 50.8 & 75.6 & 46.4 & 54.9 / 49.0 / 37.3 & 52.6 \\
    \multicolumn{1}{c|}{} & \multicolumn{1}{c|}{LateFusion} & 77.6 & 64.2 & 66.3 / 60.6 / 50.3 & 63.8 & 76.1 & 60.2 & 66.1 / 59.7 / 45.6 & 61.5 \\
    \multicolumn{1}{c|}{} & \multicolumn{1}{c|}{ASFormer} & 77.7 & \textbf{89.0} & 78.7 / 73.0 / 59.1 & 75.5 & 76.6 & \textbf{90.5} & 87.4 / 81.3 / 64.0 & 80.0 \\
    \multicolumn{1}{c|}{} & \multicolumn{1}{c|}{EmbryosFormer} & 77.4 & 71.0 & 72.6 / 69.4 / 61.5 & 70.4 & 77.0 & 81.0 & 81.3 / 75.0 / 61.6 & 75.2 \\
    \multicolumn{1}{c|}{} & \multicolumn{1}{c|}{FACT} & 78.9 & 83.6 & 79.6 / 74.5 / 65.6 & 76.4 & 76.1 & 87.0 & 86.5 / 80.8 / 65.9 & 79.3 \\
    \midrule
    
    \multicolumn{2}{c|}{EmbryoDiff (1 step)} & 82.8 & 82.8 & 81.0 / 75.9 / 65.8 & 77.7 & 81.3 & 83.5 & 85.9 / 81.1 / 68.9 & 80.1 \\
    
    \multicolumn{2}{c|}{EmbryoDiff (15 steps)} & 83.1 & 86.3 & 83.4 / 78.2 / 68.4 & 79.9 & 81.2 & 86.8 & 88.1 / 83.7 / 71.2 & 82.2 \\

    \multicolumn{2}{c|}{EmbryoDiff (25 steps)} & \textbf{83.1} & 86.7 & \textbf{83.7} / \textbf{78.6} / \textbf{68.9} & \textbf{80.2} & \textbf{81.3} & 86.8 & \textbf{88.4} / \textbf{84.2} / \textbf{71.3} & \textbf{82.4} \\

    \bottomrule
    \end{tabular}
    \caption{Comparison with the state-of-the-art methods on Multi-Focal Human Embryos Dataset and Single-Focal Human Embryos Dataset. In the table, ``Frame'' means frame-based methods and ``Sequence'' stands for sequence-based methods.}
  \label{tab:comp_sota}%
\end{table*}%

\begin{table*}[t]
  \centering
    \begin{tabular}{
      p{7em}<{\centering}|       
      p{1.5em}<{\centering}      
      p{1.5em}<{\centering}   
      p{1.5em}<{\centering}   
      p{1.5em}<{\centering}   
      p{1.5em}<{\centering}   
      p{1.5em}<{\centering}   
      p{1.5em}<{\centering}   
      p{1.5em}<{\centering}  
      p{1.5em}<{\centering}    
      p{1.5em}<{\centering}    
      p{1.5em}<{\centering}    
      p{1.5em}<{\centering}    
      p{1.5em}<{\centering}    
      p{1.5em}<{\centering}    
      p{1.5em}<{\centering}    
    }
    \toprule
    Methods & tPB2 & tPNa & tPNf & t2 & t3 & t4 & t5 & t6 & t7 & t8 & t9+ & tM & tSB & tB & tEB \\
    \midrule
    ResNet & 64.4 & 92.0 & \underline{91.7} & 90.6 & 49.3 & 75.6 & \textbf{38.2} & 27.0 & 26.7 & 65.6 & 75.3 & \underline{75.4} & 66.9 & 49.3 & 88.2 \\
    DLTEmbryo & 34.6 & 91.3 & 76.7 & 82.2 & 31.1 & 78.7 & 11.8 & 15.2 & 4.5 & 76.0 & 68.0 & 70.5 & 65.9 & 24.9 & 91.0  \\
    InceptionV3 & 63.2 & 94.7 & 89.5 & 90.2 & \textbf{55.9} & 80.5 & 26.7 & 23.1 & 27.4 & 73.5 & 80.1 & 72.5 & 65.3 & 42.9 & 90.1  \\
    ResNet-LSTM & 58.9 & 95.0 & 90.3 & 92.5 & \underline{53.4} & 78.6 & 35.1 & \underline{31.6} & \textbf{30.2} & 61.9 & 77.7 & 66.6 & 71.8 & 34.2 & \textbf{91.2}  \\
    LateFusion & \underline{91.9} & 95.1 & 91.3 & 95.1 & 47.6 & 79.1 & 36.0 & 31.2 & 34.1 & 61.0 & 78.8 & 73.2 & 72.9 & 48.4 & \underline{91.1} \\
    ASFormer & 87.6 & 96.2 & 77.5 & 90.7 & 28.6 & 86.6 & 20.5 & 28.5 & 22.5 & 78.6 & 74.3 & 73.0 & 78.9 & 37.4 & 93.4 \\
    EmbryosFormer & 86.0 & \underline{97.1} & 74.2 & 95.7 & 0.0 & \textbf{97.4} & 0.7 & 8.3 & 0.0 & \textbf{86.1} & \underline{81.1} & 58.3 & \textbf{87.2} & 2.5 & 84.9 \\
    FACT & 88.6 & 95.8 & 83.0 & \underline{96.0} & 17.2 & 87.2 & 28.3 & 22.2 & 23.2 & 75.3 & 80.1 & 75.2 & 75.2 & \underline{51.4} & 90.1\\
    \midrule
    EmbryoDiff & \textbf{93.2} & \textbf{97.2} & \textbf{93.7} & \textbf{97.0} & 51.1 & \underline{90.9} & \underline{36.0} & \textbf{35.0} & \underline{28.5} & \underline{83.4} & \textbf{81.8} & \textbf{79.0} & \underline{84.9} & \textbf{52.7} & 90.4 \\
    \bottomrule
  \end{tabular}
  \caption{Per-class accuracy comparison on MFHE Dataset. For EmbryoDiff, we use 25 denoising steps.}
  \label{tab:per_class_comp}
\end{table*}

\section{Experiments}
\subsection{Datasets}

\subsubsection{Multi-Focal Human Embryos Dataset (MFHE)}
We use the MFHE dataset~\cite{dataset_gomez} to evaluate the effectiveness of multi-focal feature fusion and the proposed diffusion framework. It contains 704 embryo sequences with 7 focal planes and 16 developmental stages. We refine the annotations and remove the rare tHB class, resulting in 655 videos. The dataset is split 7:3 into train and test sets.

\subsubsection{Single-Focal Human Embryos Dataset (SFHE)}
We use the SFHE dataset~\cite{seq_embryosformer} to assess the generality and single-focal adaptability of our framework. It contains 440 embryo sequences with single-focal videos and 12-stage labels. We use the original train set for training and merge the validation and test sets for testing.



\subsection{Evaluation Metrics} 
To evaluate temporal consistency and boundary localization accuracy, we use standard metrics from action segmentation: Accuracy, Edit Score, and F1@{10, 25, 50}. While Accuracy measures frame-wise correctness, Edit Score and F1 scores assess temporal segment quality, offering a more comprehensive evaluation. 

\subsection{Implementation Details}
In Stage 1, a ResNet-50 \cite{resnet} backbone is trained using the AdamW optimizer, a batch size of 32, and an initial learning rate of $1 \times 10^{-4}$ with cosine decay over 20 epochs. In Stage 2, frame-wise features with dimension $D = 2048$ are extracted from all focal planes. Each condition encoder consists of 6 layers with a reduced hidden dimension of 96. Intermediate features from $\{2, 4, 6\}$ layers are concatenated to form the condition signals. The diffusion model consists of 8 hybrid conditioning blocks with a feature dimension of 128. We follow the cosine noise schedule with 1000 diffusion steps and set the input signal scale to 0.1 The model is trained for 350 epochs with a batch size of 24, an initial learning rate of $1 \times 10^{-4}$, and a weight decay of 0.01 using AdamW and cosine learning rate decay.  All models are trained on the training set using a single NVIDIA RTX 4090 GPU and the best evaluation results on test set are reported. For more details about the dataset and experimental setups, please see the extended version.

\subsection{Comparison with SOTAs}
We compare EmbryoDiff with frame-based (e.g., ResNet \cite{dataset_gomez}), sequence-based (e.g., EmbryosFormer \cite{seq_embryosformer}), and top-performing action segmentation models (ASFormer \cite{ASFormer}, FACT \cite{FACT}). Open-source methods are evaluated using official codes; closed-source ones (DLTEmbryo \cite{frame_DTLEmbryo}, LateFusion \cite{seq_latefusion}) are carefully reimplemented. For our model, we report average results over ten runs with different random seeds.

\subsubsection{Overall Performance Comparison}
Detailed results are listed in Table \ref{tab:comp_sota}. On MFHE dataset, frame-based methods such as InceptionV3 achieve reasonable frame-level accuracy (76.2\%) but suffer in sequence-level metrics (F1@50: 23.9, Edit: 29.8) due to lack of temporal modeling. In contrast, sequence-based models like FACT, which uses complex designs to model temporal dependencies, achieve better accuracy (78.9\%) and significantly improved F1@50 (65.6\%). However, they still struggle with challenging samples, likely due to their discriminative nature and reliance on single-focal videos. In comparison, our method outperforms all baseline methods on the average metric with only a single denoising step. As the denoising steps increases to 25, EmbryoDiff achieves 83.1\% Accuracy and 68.9\% F1@50, setting a new state of the art. While ASFormer performs well on the Edit score (89.0), it lags behind our model on other metrics, particularly F1@50 (59.1\%), indicating its limited ability to precisely localize stage boundaries.

On SFHE dataset, where all methods use single-focal input, our EmbryoDiff approach still achieves the best average metric using only one denoising step, demonstrating the superiority of our diffusion-based framework. With 25 sampling steps, our method reaches 81.3\% Accuracy, improving upon the second-best method, EmbryosFormer, by 5.58\%. Moreover, our F1@50 reaches 71.3\%, surpassing the second-best FACT (65.9\%) by 8.19\%.

\begin{figure}[t]
\centering
\includegraphics[width=0.85\columnwidth]{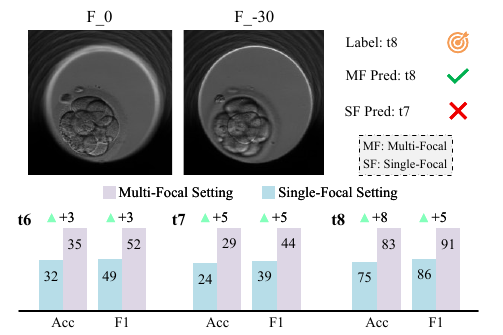} 
\caption{Advantages of multi-focal feature fusion over single-focal. The top shows it better handles cell occlusions, and the bottom shows it greatly improves late cleavage stage identification.}
\label{fig:MFvsSF}
\end{figure}

\subsubsection{Per-Class Accuracy Comparison}
In Table~\ref{tab:per_class_comp}, we present a per-class accuracy comparison across methods on MFHE dataset. Our proposed EmbryoDiff achieves the best or second-best accuracy on the majority of classes, demonstrating its strong discriminative capability across diverse developmental stages. Notably, for the multi-cell stages (t6, t7, t8, t9+), EmbryoDiff consistently delivers top-tier performance. In contrast, while some methods may excel on one stage, they often suffer from severe performance drops on others. For example, EmbryosFormer achieves 86.1\% accuracy on t8, but performs poorly on other stages, with only 8.3\% on t6 and 0.0\% on t7. The results highlight the superiority of our method in fine-grained embryo stage recognition.

\subsection{Ablation Studies}

\subsubsection{Effectiveness of Key Designs}

In Table \ref{tab:ablation_mf}, we ablate key components of EmbryoDiff on MFHE dataset. The designs include Multi-Focal Feature Fusion (MFF), complementary Semantic Condition Encoder (SCE) and Boundary Condition Encoder (BCE), and the Conditional Diffusion Model (CDM). Specifically, comparing Rows 1 and 2 as well as Rows 3 and 4 reveals that both the discriminative model (SCE only) and the diffusion-based generative model benefit from MFF, with the latter showing a more significant improvement. This highlights the enhanced capacity of the diffusion framework to leverage multi-focal information. Additionally, the diffusion-based framework outperforms the semantic encoder-only architecture (Row 1 vs. Row 3 and Row 2 vs. Row 4), demonstrating its strength in leveraging the distribution priors of embryo development to improve the predictions. Finally, the comparison across Rows 4–6 confirms that both the proposed complementary BCE and MFF contribute substantially to performance gains. Each component consistently improves prediction accuracy and sequence-level metrics, confirming our statements.

The ablation results on the SFHE dataset (Table~\ref{tab:ablation_sf}) exhibit similar trends. The conditional diffusion model significantly outperforms the baseline that uses only semantic condition encoder. By incorporating boundary condition features through the proposed Hybrid Semantic-Boundary Condition Block, the prediction performance improves further, particularly on sequence-level metrics such as Edit Score and F1@50. This indicates that boundary-aware signals help the diffusion model better localize transitional frames in the developmental sequence.

\begin{table}[tbp]
  \centering
    \begin{tabular}{
      p{2.0em}<{\centering}  
      p{2.0em}<{\centering}  
      p{2.0em}<{\centering}  
      p{2.0em}<{\centering}|  
      p{2.2em}<{\centering}  
      p{2.2em}<{\centering}  
      p{2.8em}<{\centering}  
    }
    \toprule
    SCE & CDM & BCE & MFF & Acc & Edit & F1@50 \\
    \midrule
    \checkmark  &   &   &   & 80.1 & 78.3 & 60.2 \\  
    \checkmark &   &   & \checkmark  & 81.3 & 81.2 & 63.0 \\  

    \checkmark & \checkmark &   &   & 81.0 & 82.8 & 62.2 \\  

    \checkmark & \checkmark & & \checkmark  & 82.5 & 85.3 & 66.0 \\  

    \checkmark & \checkmark & \checkmark & & 80.6 & 83.3 & 63.9 \\  

    \checkmark & \checkmark & \checkmark & \checkmark & \textbf{83.1} & \textbf{86.3} & \textbf{68.4} \\  
    \bottomrule
    \end{tabular}
    \caption{Ablation study results of key designs on MFHE dataset using 15 denoising steps.}
  \label{tab:ablation_mf}%
\end{table}

\begin{table}[t]
  \centering
    \begin{tabular}{
      p{2.0em}<{\centering}  
      p{2.0em}<{\centering}  
      p{2.0em}<{\centering}|  
      p{2.3em}<{\centering}  
      p{2.3em}<{\centering}  
      p{2.8em}<{\centering}  
    }
    \toprule
    SCE & CDM & BCE & Acc & Edit & F1@50 \\
    \midrule

    \checkmark & &  & 80.5 & 81.8 & 67.5 \\  

    \checkmark & \checkmark & & 80.8 & 85.8 & 70.2 \\  

    \checkmark & \checkmark & \checkmark & \textbf{81.2} & \textbf{86.8} & \textbf{71.2} \\  
    \bottomrule
    \end{tabular}
    \caption{Ablation study results of key designs on SFHE dataset using 15 denoising steps.}
  \label{tab:ablation_sf}%
\end{table}

\begin{figure*}[tpb]
\centering
\includegraphics[width=0.95\textwidth]{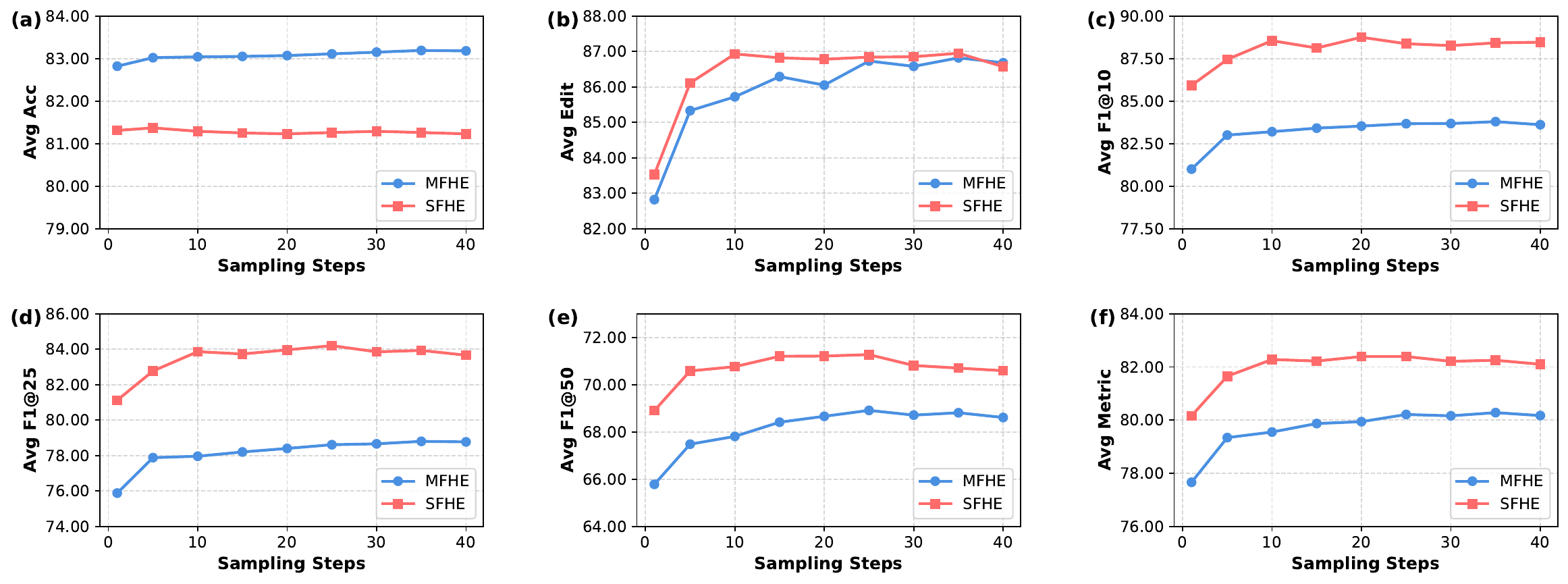} 
\caption{The trends of evaluation metrics on two datasets (MFHE and SFHE) with respect to denoising sampling steps during inference. From (a) to (f), the metrics are Accuracy, Edict Score, F1@\{10, 25, 50\}, and Average Metric, respectively.}
\label{fig:metric_over_steps}
\end{figure*}

\subsubsection{Superiority of Multi-Focal Feature Fusion}


To highlight the critical role of multi-focal feature fusion in the task of fine-grained embryo developmental stage recognition, we perform an in-depth analysis in Figure~\ref{fig:MFvsSF}. As shown in the upper part, the central focal plane image (left) clearly reveals seven cells, along with a faint, out-of-focus contour beneath. Relying solely on a single focal plane may lead to misinterpretation of cell count and, consequently, incorrect stage classification. In contrast, the TLM image at a deeper focal plane (right) unambiguously reveals an additional large cell, confirming its presence. By integrating information across multiple focal planes, our method reconstructs a more accurate 3D morphological structure, enabling correct developmental stage assignment.

In the lower part of Figure~\ref{fig:MFvsSF}, we compare prediction results for late cleavage multi-cellular stages under single-focal and multi-focal settings. These stages are particularly challenging due to frequent cell occlusions. As shown, multi-focal fusion yields significant performance gains in the t6, t7, and t8 stages compared to the single-focal counterpart. Moreover, the advantage becomes increasingly pronounced as the cell count increases, demonstrating its effectiveness in resolving structural ambiguities in densely packed and occluded configurations.

\begin{table}[tp]
  \centering
  \begin{tabular}{
    p{3em}<{\centering}|  
    p{2.5em}<{\centering}  
    p{2.5em}<{\centering}  
    p{7em}<{\centering}  
    p{2.5em}<{\centering}  
  }
  \toprule
  Planes & Acc & Edit & F1@\{10, 25, 50\} & Avg \\
  \midrule
  1      & 80.6 & 83.3 & 80.4/74.9/63.9 & 76.6 \\  
  3      & 81.4 & 84.8 & 81.9/76.7/66.1 & 78.2 \\  
  5     & 82.5 & 86.2 & 82.7/77.8/67.9 & 79.4 \\  
  7      & \textbf{83.1} & \textbf{86.3} & \textbf{83.4/78.2/68.4} & \textbf{79.9} \\  
  \bottomrule
  \end{tabular}
  \caption{Ablation study results of the number of focal planes used with 15 denoising steps. We symmetrically select focal planes around the central focal plane.}
  \label{tab:focal_planes}
\end{table}

\subsubsection{Effect of Focal Plane Numbers}
In Table~\ref{tab:focal_planes}, we investigate the impact of fusing varying numbers of focal planes on model performance. It is evident that prediction accuracy improves significantly as more focal plane information is incorporated. This indicates that multi-focal fusion helps the model better capture 3D embryonic structure and improve recognition performance.

\subsubsection{Effect of Denoising Steps}
We further analyze the impact of inference-time denoising steps on model performance. As shown in Figure~\ref{fig:metric_over_steps}, the sequence-level metrics (b–f) improve markedly as the number of diffusion steps increases, suggesting that the model progressively refines its predictions by leveraging learned developmental distribution priors. Performance plateaus after about 20 steps. In contrast, accuracy (a) shows only marginal gains. We hypothesize that it is limited by the bottleneck of the condition features.

\subsection{Qualitative Comparison}
\label{sec: qq}
In Figure 6, we visualize predicted developmental stage sequences from several leading models. Frame-based methods (e.g., InceptionV3), which lack explicit temporal modeling, produce temporally inconsistent predictions. Sequence-based approaches achieve better temporal coherence but still exhibit significant classification errors and inaccurate boundary localization. In contrast, our model achieves more accurate boundary detection and generates sequences that are more consistent with embryonic development patterns, demonstrating its superiority. Additional results are provided in the extended online version.

\begin{figure}[t]
\centering
\includegraphics[width=0.85\columnwidth]{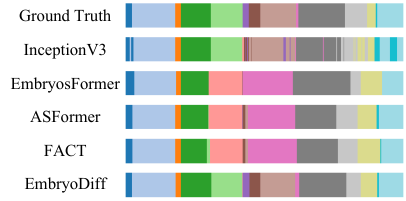} 
\caption{An example of qualitative comparison of developmental stage predictions on the MFHE dataset.}
\label{fig:preds}
\end{figure}

\section{Conclusion}
In this paper, we propose EmbryoDiff, the first diffusion-based framework for embryonic stage recognition, introducing a novel two-stage paradigm that leverages conditional diffusion models to generate accurate and temporally coherent stage sequences. By incorporating TLM video features as conditional inputs, our model guides the denoising process to recover clean stage labels from noisy predictions. To address the issue of incomplete morphological representation, we innovatively fuse multi-focal-plane TLM features, capturing a more holistic view of the 3D embryo structure. Furthermore, to provide richer conditional guidance, we extract complementary semantic and boundary cues from the fused sequence features and design a Hybrid Semantic-Boundary Condition Block to effectively integrate them into the diffusion process. Finally, we improve the annotation quality of an existing dataset and adopt more comprehensive evaluation metrics. Extensive experiments and visualizations demonstrate the superiority of our method. 

\bibliography{EmbryoDiff}

\input{Appendix/TechnicalAppendix}

\end{document}

%% file: Appendix/TechnicalAppendix.tex






\begin{figure*}[htpb]
\centering
\includegraphics[width=0.90\textwidth]{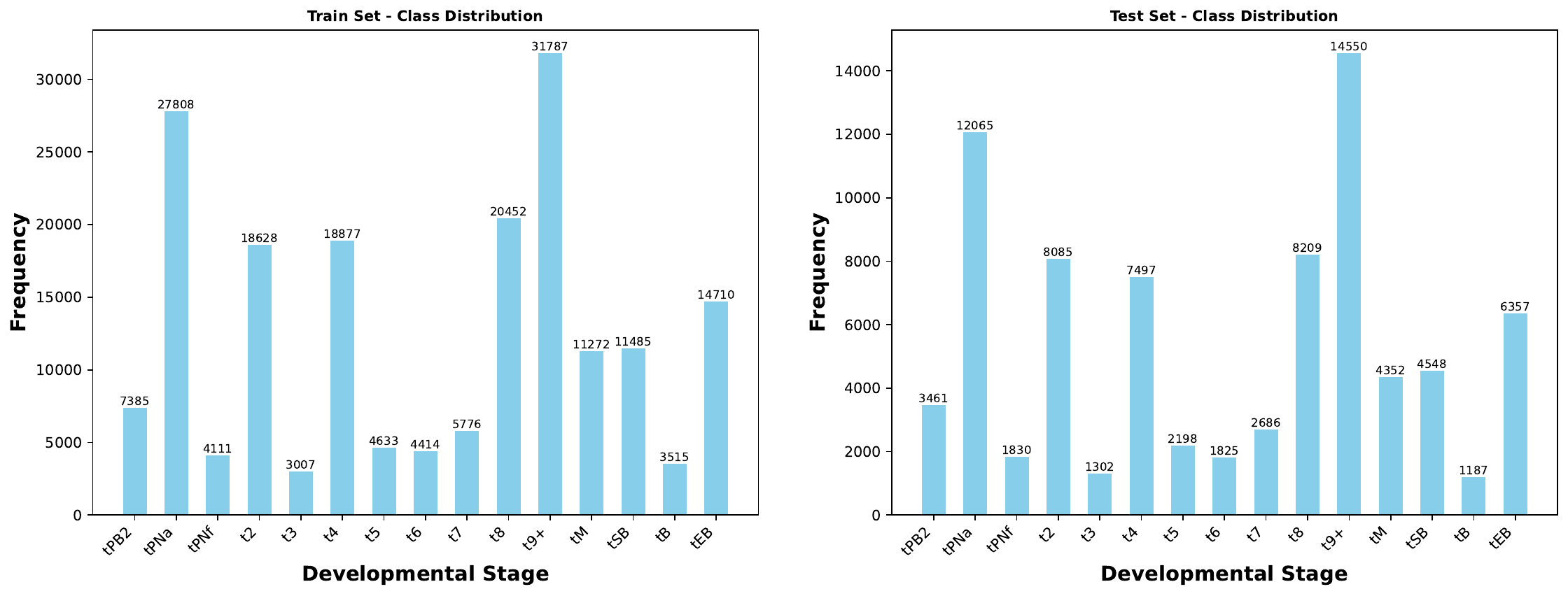} 
\caption{Class distributions for the MFHE dataset on training and test sets.}
\label{fig:distribution}
\end{figure*}

\section{Dataset Details}
\subsection{Multi-Focal Human Embryos Dataset (MFHE)}
The dataset was released by the University Hospital of Nantes, France~\cite{dataset_gomez}, and contains 704 embryo sequences from intracytoplasmic sperm injection (ICSI). Each embryo is represented as a multi-focal time-lapse monitoring (TLM) video with seven focal planes. The dataset includes 16 developmental stage labels: tPB2, tPNa, tPNf, t2, t3, t4, t5, t6, t7, t8, t9+, tM, tSB, tB, tEB, and tHB. Due to label noise~\cite{dataset_problem}, we improve the annotation quality by removing incorrect annotations and refining boundary labels. Following~\cite{seq_canat2024novel}, we exclude the tHB category, which has very few samples. The final dataset contains 655 embryo TLM videos, randomly split into training and test sets in a 7:3 ratio. As shown in Figure~\ref{fig:distribution}, the class distribution is consistent between the training and test sets, ensuring a fair and representative evaluation.

\subsubsection{Annotation Correction}
We provide further details on the annotation refinement process. The annotation issues are categorized into three main types:

\begin{itemize}
    \item \textbf{Dark Video Frames:} Some videos contain prolonged sequences of dark or severely out-of-focus frames that are nonetheless annotated with developmental stages. These videos are entirely removed to prevent introducing misleading signals during feature learning. 
    \item \textbf{Incorrect Stage Labels:} In certain sequences, frames are mislabeled (e.g., t2 stages incorrectly labeled as t4). These errors are corrected through frame-by-frame review conducted by experienced annotators.
    \item \textbf{Inaccurate Boundary Annotations:} While the general sequence of developmental stages is correctly identified in certain instances, the temporal precision of stage transitions frequently falls short. Such inaccuracies impede the model's capacity to extract robust and discriminative features. To address this, we meticulously refine the annotations around stage transition boundaries by performing manual, frame-by-frame corrections.
\end{itemize}

These corrections significantly improve the dataset quality. The refined dataset will be made publicly available upon acceptance.

\subsection{Sing-Focal Human Embryos Dataset (SFHE)} 
This dataset was introduced by EmbryosFormer~\cite{seq_embryosformer} and contains 440 ICSI embryo sequences from 112 patients, with images captured every 15+ minutes at approximately $400 \times 400$ resolution (average 339 frames per sequence). The sequences are annotated by three embryologists with both coarse (4-stage) and fine-grained (12-stage) labels. The 12 fine-grained stages are: tPNa, tPNf, t2, t3, t4, t5, t6, t7, t8, tSC, tM, tSB. In this paper, we use the fine-grained annotations for detailed stage recognition. Since the dataset provides only single-focal-plane TLM videos, we disable the Multi-Focal Feature Fusion component in EmbryoDiff. We merge the original validation and test sets into a unified test set and adopt the data preprocessing and transformation pipeline used by EmbryosFormer\cite{seq_embryosformer}.

\begin{table*}[htpb]
  \centering
    \begin{tabular}{
      p{7em}<{\centering}|
      p{1.5em}<{\centering} 
      p{1.5em}<{\centering}
      p{1.5em}<{\centering}   
      p{1.5em}<{\centering}
      p{1.5em}<{\centering}  
      p{1.5em}<{\centering} 
      p{1.5em}<{\centering}  
      p{1.5em}<{\centering}   
      p{1.5em}<{\centering}   
      p{1.5em}<{\centering}  
      p{1.5em}<{\centering}   
      p{1.5em}<{\centering}   
    }
    \toprule
    Methods & tPNa & tPNf & t2 & t3 & t4 & t5 & t8 & tSC & tM & tSB & tB & tHB \\
    \midrule
    ResNet & 98.9 & 77.9 & 91.9 & 39.5 & 85.1 & 38.8 & 72.6 & 67.2 & 61.8 & 77.5 & 71.1 & 34.3 \\
    DLTEmbryo & 98.9 & 57.2 & 87.9 & 15.2 & 80.4 & 28.6 & \underline{75.8} & 56.2 & 59.8 & 82.2 & 60.8 & 17.3  \\
    InceptionV3 & 98.6 & 76.0 & 90.6 & 39.8 & 80.4 & 45.6 & 82.4 & 54.5 & 67.3 & 79.4 & 72.3 & 33.4  \\
    ResNet-LSTM & 97.6 & 81.0 & 89.6 & \underline{44.1} & \underline{86.3} & \underline{49.5} & 75.7 & 55.6 & 68.4 & 82.4 & 59.0 & 42.7  \\
    LateFusion & 98.8 & \textbf{81.9} & 88.5 & 42.7 & 76.7 & 45.8 & \textbf{76.3} & \underline{68.4} & 63.9 & \textbf{84.4} & 65.1 & \textbf{47.0} \\
    ASFormer & 98.1 & 75.4 & 91.5 & 38.4 & 83.3 & 36.2 & 71.3 & 64.7 & 77.5 & 73.8 & 73.3 & 39.3 \\
    EmbryosFormer & \underline{99.3} & 78.1 & \textbf{95.7} & 0.0 & 90.0 & 24.2 & 62.9 & 59.0 & \textbf{85.1} & 73.1 & \textbf{77.5} & 0.0 \\
    FACT & 98.5 & 75.1 & 92.2 & 30.0 & \textbf{87.4} & 28.7 & 72.2 & 64.9 & 74.4 & 74.1 & 66.9 & 43.4 \\
    \midrule
    EmbryoDiff & \textbf{99.5} & \underline{81.3} & \underline{93.7} & \textbf{58.2} & 84.1 & \textbf{57.4} & 73.7 & \textbf{72.5} & \underline{80.1} & \underline{82.8} & \underline{75.1} & \underline{46.8} \\
    \bottomrule
  \end{tabular}
  \caption{Per-class accuracy comparison on SFHE Dataset. For EmbryoDiff, we use 25 denoising steps.}
  \label{tab:per_class_comp_sf}
\end{table*}

\section{Experiments}

\subsection{Implementation Details}

We provide additional details regarding the implementation of baseline methods and the training configurations used in our experiments.

For methods including ResNet~\cite{dataset_gomez}, InceptionV3~\cite{frame_inception}, DTLEmbryo~\cite{frame_DTLEmbryo}, ResNet-LSTM~\cite{dataset_gomez}, and LateFusion~\cite{seq_latefusion}, we implement the models exactly as described in their original papers, following the reported architectures. All of these models are trained under a consistent configuration until full convergence: a batch size of 32, an initial learning rate of $1 \times 10^{-4}$, decayed to a minimum of $1 \times 10^{-7}$ using a cosine learning rate schedule, a weight decay of 0.01, and a total of 20 training epochs. For ASFormer~\cite{ASFormer}, EmbryosFormer~\cite{seq_embryosformer}, and FACT~\cite{FACT}, we use their official codebases and the recommended hyperparameters for training. All models are trained until convergence.

For data preprocessing, we adopt dataset-specific protocols. On the MFHE dataset, we recompute the mean and standard deviation across the entire dataset for input normalization. During training, images are resized to $256 \times 256$, followed by random cropping to $224 \times 224$, and augmented with random horizontal and vertical flips, rotation, and ColorJitter to improve robustness to illumination and pose variations. At test time, images are resized to $256 \times 256$ and center-cropped to $224 \times 224$ without additional augmentation. For the SFHE dataset, we follow the preprocessing and training protocol from the original EmbryosFormer \cite{seq_embryosformer} to ensure a fair comparison.

\begin{figure*}[htpb]
\centering
\includegraphics[width=0.95\textwidth]{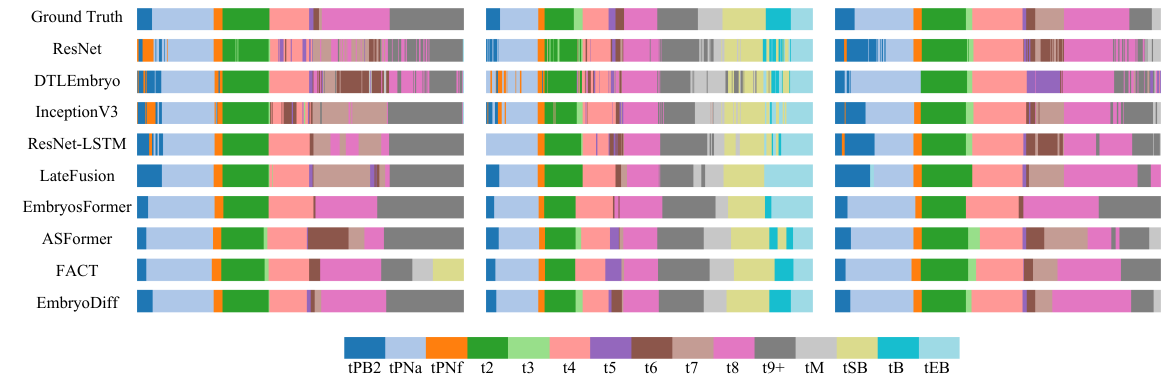} 
\caption{Additional qualitative comparisons of embryo developmental stage classification results on the MFHE dataset. The classification results of our EmbryoDiff are obtained using 15 denoising steps.}
\label{fig:MFHE_preds}
\end{figure*}

\begin{figure*}[htpb]
\centering
\includegraphics[width=0.95\textwidth]{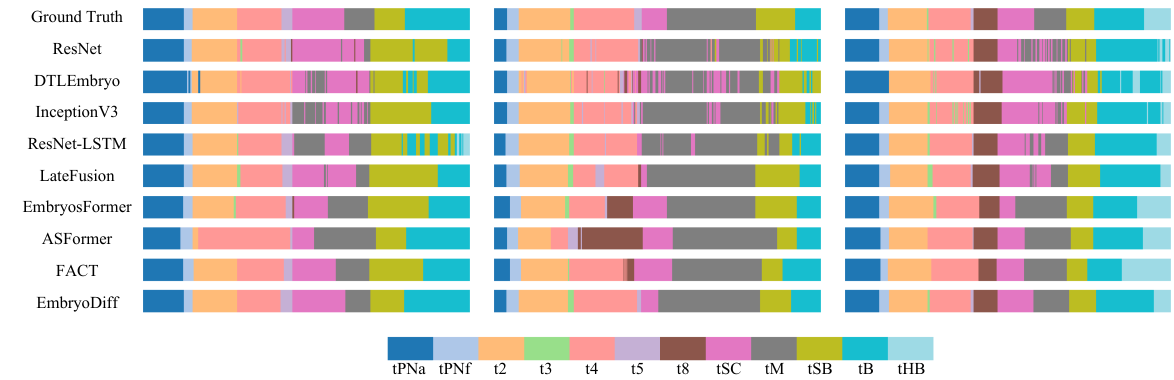} 
\caption{Additional qualitative comparisons of embryo developmental stage classification results on the SFHE dataset. The classification results of our EmbryoDiff are obtained using 15 denoising steps.}
\label{fig:SFHE_preds}
\end{figure*}

\subsection{Additional Experiments}
\subsubsection{Per-Class Accuracy Comparison on SFHE Dataset}
In Table~\ref{tab:per_class_comp_sf}, we further report the per-class accuracy of each method on the SFHE dataset. As shown, even using single-focal-plane TLM videos as input, our EmbryoDiff achieves the best or second-best performance in 10 out of the 12 classes. This clearly demonstrates the superiority of our diffusion-based approach in recognizing diverse developmental stages compared to other methods.

\subsubsection{More Qualitative Comparisons}
In Figure~\ref{fig:MFHE_preds}, we present additional representative developmental stage classification results of various methods on the MFHE dataset. Frame-based approaches, particularly ResNet and DTLEmbryo, exhibit poor temporal coherence and frequently violate the biological principle of monotonic stage progression. Although InceptionV3 achieves relatively better performance within this group, it still produces noticeable timing inconsistencies, especially in morphologically similar stages such as t9+ and tM, or tSB and tB, where appearance changes are subtle and single-frame predictions are inherently ambiguous. Sequence-based methods address this limitation by modeling stage transitions over time, resulting in more temporally consistent predictions and largely avoiding invalid stage orders. However, they still suffer from imprecise boundary localization, indicating that visual features alone are insufficient for fine-grained discrimination of closely spaced developmental phases. In contrast, our diffusion-based framework leverages distributional priors such as typical stage durations and temporal co-occurrence patterns to improve the prediction accuracy. As shown in Figure~\ref{fig:MFHE_preds}, our proposed EmbryoDiff produces more accurate and temporally plausible predictions, demonstrating its superiority over comparing methods.

Figure~\ref{fig:SFHE_preds} shows qualitative comparisons on the SFHE dataset. The observed trends closely mirror those on MFHE, confirming the generalization of the findings. Notably, our method achieves accurate predictions even for rare stages such as t3 and t5, where limited training samples make learning challenging for conventional models. By leveraging the underlying embryo development distributional priors, our proposed EmbryoDiff effectively regularizes predictions and maintains high accuracy across diverse and imbalanced developmental stage distributions.

